\documentclass{IEEEtran}
\usepackage{bm}
\usepackage{amsmath}
\usepackage{amssymb}
\usepackage{booktabs}
\usepackage{algorithm}
\usepackage{algorithmicx}
\usepackage{algpseudocode}
\usepackage{stfloats}
\usepackage{graphicx}
\usepackage{subfigure}
\usepackage{pdfpages}
\usepackage{graphicx}
\usepackage{wrapfig}
\usepackage{listings}
\usepackage{color}
\usepackage[colorlinks,linkcolor=black,anchorcolor=blue,citecolor=blue]{hyperref}
\usepackage{alltt}
\usepackage{marvosym}
\usepackage{xcolor}
\input{highlight.sty}

\title{Characters Detection on Namecard with faster RCNN}
\author{
    Weitong~ZHANG
    \thanks{W.~ZHANG Email \texttt{WeightZero@outlook.com}}
}
\begin{document}
\maketitle
\begin{abstract}
    We apply Faster R-CNN to the detection of characters in namecard, in order to 
    solve the problem of a small amount of data and the inbalance between different class,
    we designed the data augmentation and the 'fake' data generalizer to generate more data
    for the training of network. Without using data augmentation, 
    the average IoU in correct samples could be no less than 80\% and the mAP result of 80\% 
    was also achieved with Faster R-CNN. By applying the data augmentation, the variance of mAP
    is decreased and both of the IoU and mAP score has increased a little.
\end{abstract}
\section{Introduction}
Nowadays, there are lots of successful solutions to the machine translation problem, such as the
Convolutional Sequence to Sequence Learning\cite{GehringAGYD17} method and the Attention method \cite{VaswaniSPUJGKP17}.
Researchers are even using some method to speed up the training\cite{DBLP:journals/corr/abs-1709-02755}
and the inference period\cite{DBLP:journals/corr/GuCL17}. However, in a more sophisticated case, the machine translator
would be asked to translate many kinds of language at the same time, e.g. in a picture. Therefore, it is also a problem
to detect the different kinds of language in a picture and convert them into text. As a simple trial of this problem,
we have collected about 1,500 namecards and use the faster rcnn method to detect the Chinese characters, English characters
and numbers.

The remaining of this paper is organized as follows. Section 2 describes the basic structure of Faster R-CNN, including the 
training method and testing method. Section 3 introduce the data augmentation and generation of namecard. 
The result of the experiment and a brief analysis are presented in Section 4. Finally, concluding remarks are presented in Section 5, while the brief introduction of the code would be in Section 6.

\section{Overview of the faster RCNN}
\subsection{Basic structure of faster R-CNN}
\begin{figure}[h]
    \centering
    \includegraphics[width=\columnwidth]{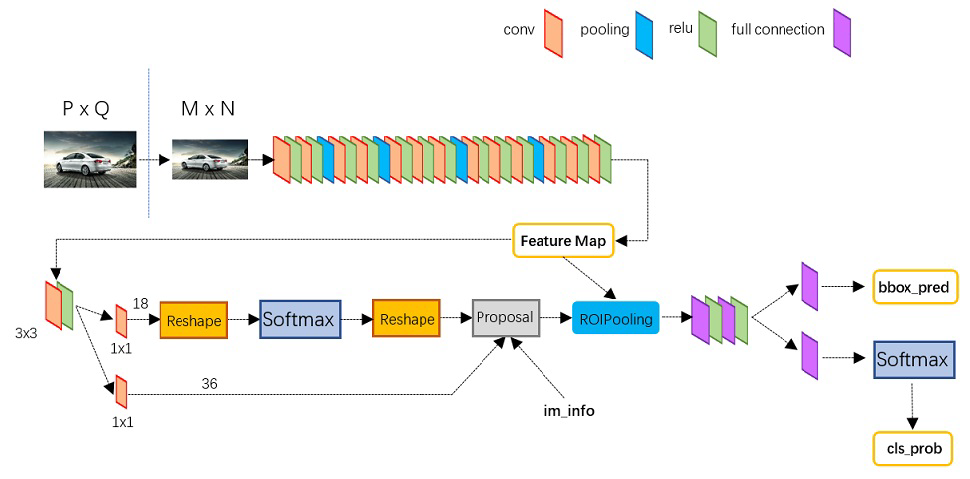}
    \caption{Network structure}
    \label{FasterRCNN}
\end{figure}

Faster R-CNN\cite{renNIPS15fasterrcnn}, proposed by Ren et al, relies on a two-stage object detection, as Fig. \ref{FasterRCNN} presents. First, a sub-network is used to propose the bounding boxes, second, a separate sub-network is used to classify the objects within each box.

\subsubsection{Region Proposal Network (RPN)}
The first stage is an RPN network. For each picture input, a pre-trained network (usually trained on ImageNet 
ILSVRC15\cite{ILSVRC15}) is used to exact the feature of the picture and generate a 'feature map'. The structure of 
the pre-train network could be VGG\cite{VGG} or ResNet\cite{RES}. In Fig.1, we use VGG16 as the feature exactor 
while in the experiment, ResNet101 is used due to a better performance。

For each 'sup-pixel' in the exacted feature map, we can locate a pixel in the original picture which is the center 
of the receptive field 
of the 'sup-pixel'. Therefore, we can set up a series of 'anchor box',whose center is that certain pixel in the 
original pircure as the 
guess of the bounding box. For each 'sup-pixel' in the feature map, we can generate $k$ anchors box with different 
ratio and size. 
Therefore, suppose that we have a $m \times n$ sized feature map, we can finally got $k * m * n$ anchors.

The feature map is connected with a $3\times 3$ convolutional layer and the output is separately connected with two 
$1 times 1$ sibling convolutional layer. The output of the first convolutional layer (the upper one in Fig. \ref
{FasterRCNN}) is a map with a $2k$-d sup-pixel, indicating the $k$-th anchor generated from this position (the 
sup-pixel of the feature map in the same position, to be precise) is a bounding box or not. While the output of the 
second convolutional layer (the lower one in Fig. \ref{FasterRCNN}) is a map with a $4k$-d sup-pixel, indicating 
the correction of each anchors and generate a more precise bounding box.

In pratice, we choose $k = 9$ just like the original faster R-CNN did.

\subsubsection{Classifier network}
The second stage is a classifier, which contains a RoI pooling layer proposed by Ross Girshick in Fast RCNN\cite
{girshickICCV15fastrcnn}. 
The classifier is also working with the feature map generated from the feature exactor mentioned in RPN network. 
First of all, from the 
relationship between the pixel of the feature map and the pixel of the original picture, we can transform the 
bounding box in the 

original map (generated by RPN module) into a bounding box in the feature map.

The cropped feature map (suppose its size is $m \times n$) is fed into the RoI pooling layer and generated a 
$7\times7$ output tensor. 
The RoI pooling rule is that the pooling step is calculated according to the size of the cropped feature map. In 
our case, the pooling 
step is $(m/7,n/7)$ to generate a $7\times7$ output tensor. Therefore, no matter what size of the feature map is 
fed in, the RoI pooling 
layer will definitely generate a fixed size output. This output is connected with two sibling fully connect layer, 
which is the 
regression of the bounding box and the confidence of each classes.

\subsection{Training}
We used the '4-Step Alternating Training' to train the RPN network and classifier in turn. The training logic is
\begin{itemize}
    \item {Train the RPN initialized with an ImageNet-pre-trained model and ﬁne-tuned end-to-end for the region proposal task}
    \item {Train a separate detection network by Fast R-CNN using the proposals generated by the step-1 RPN}
    \item {Use the detector network to initialize RPN training, but ﬁx the shared convolutional layers and only ﬁne-tune the layers unique to RPN}
    \item {Keep the shared convolutional layers ﬁxed, we ﬁne-tune the unique layers of Fast R-CNN.}
\end{itemize}
\subsubsection{Training on RPN}
We use SGD method to train the RPN method, the fixed parameters or layers are metioned above, the loss function is that

\begin{equation}
    L({p_i,t_i}) = \frac1{N_{cls}}\sum_iL_{cls}(p_i,p_i^*) + \frac\lambda{N_{reg}}p_i^*\sum_iL_{reg}(t_i,t_i^*)
\end{equation}
\begin{equation}
    L_{cls}(p_i,p_i^*) = p_i^*\log(p_i) \text{ (Cross Entropy)}
\end{equation}
\begin{equation}
    L_{reg}(t_i,t_i^*) = \begin{cases}
        |t_i - t_i^*|  \text{ if } |t_i - t_i^*| \le 1\\
        (t_i - t_i^*)^2 \text{ otherwise}
    \end{cases}
    \text{ (smooth L1)}
\end{equation}
\begin{equation}
    \begin{cases}
        t_x = \frac{x-x_a}{w_a}\\
        t_y = \frac{y-y_a}{h_a}\\
        t_w = \log(w / w_a) \\
        t_h = \log(h / h_a) \\
    \end{cases}
    \begin{cases}
        t_x^* = \frac{x^*-x_a}{w_a}\\
        t_y^* = \frac{y^*-y_a}{h_a}\\
        t_w^* = \log(w^* / w_a) \\
        t_h^* = \log(h^* / h_a) \\
    \end{cases}
\end{equation}
The symbols are described as the following table Tab. \ref{tableRPN}
\begin{table}[h]
    \centering\caption{symbols in RPN's loss function}\label{tableRPN}
       \begin{tabular}{cc}
        \toprule
        Symbols & Meaning \\
        \midrule
        $i$ & Index of an anchor in a mini-batch \\
        $p_i$ & Predicted probability of anchor $i$ \\
        $p^∗_i$ & 1 if the anchor is positive, otherwise 0\\
        $t_i$ & Vector represents predicted bounding box\\
        $t_i^*$ & Vector represents ground true bounding box\\
        $N_{cls}$ & Number of ground true boxes\\
        $N_{reg}$ & Number of anchors\\
        $x,y,w,h$ & Top Left corner and width,height of the predict box\\
        $x_a,y_a,w_a,h_a$ & Top Left corner and width,height of the anchor box\\
        $x^*,y^*,w^*,h^*$ & Top Left corner and width,height of the ground truth box\\
        \bottomrule
    \end{tabular}
\end{table}

According to the designation of faster RCNN, an anchor would be a positive anchor if it has $IoU > 0.8$ or it has the highest IoU among
all of the anchors.

\subsubsection{Training on Fast RCNN model}
The second part is to trian the fast RCNN model, the loss function is the same with the RPN model, the only little difference is that
in the fast RCNN model, we use the 'original predict box' (i.e. box generated by RPN model) to replace the 'anchor' box in RPN module.
Therefore, the predicted box is refined with the fast RCNN model comparing to the RPN model.

\subsection{Testing}
Since the faster RCNN is an end-to-end network, it is extremely easy to test the network, with the testing image input, the predict postive anchor is corrected (regress) by the RPN module and feed to the fast RCNN module, then the confidence score is given and the
bounding box is regressed again. Bounding box with confidence greater than 0.8 will be output.

There are several score could be used to evaluate the performance of the network.

\subsubsection{IoU score}

Set the predict bounding box is A and the ground truth bounding box is B, then the IoU score between A and B could be described as

\begin{equation}
    IoU(A,B) \triangleq \frac{A\cap B}{A\cup B}
\end{equation}

Normally, a predicted bounding box is discriminated to be a correct bounding box if the max IoU score with the ground truth box which is 
the same label with the predicted bounding box is greater than 0.8. Also, the average IoU score of all of the correct bounding box is an important score of detection task, since it is indicating the 'accurate' of the detection task.

\subsubsection{mAP score}

mAP score is the mean Average Precision score, for each image, we can calculate the precision score of each class, and for the entire 
dataset, we can calculate the Average Precision of each class. And the mean Average Precision of all class is the mAP score, the mAP 
score is the most import score to measure the performance of the model. However, mAP score might be influence by the inbalanced data.
Therefore, give an analysis of each AP is also important.

\subsubsection{F-1 score}
F-1 score is also used to calculate the preformance of each image, and we can also calculate the mean average F-1 score. The F-1 score
would be described as the following method. It is a type of mean of the precision and recall ratio

\begin{equation}
    F1 \triangleq \frac{2 * N_{accurate}}{N_{report}* N_{groud\ truth}}
\end{equation}

\section{Data Augmentation}
\subsection{Data Augmentation}
We have original 1238 namecards, carefully annotated. However, 1238 images is such a small data set for deep learning task. Therefore, we consider some data Augmentation method. We total use these image processing method provided by PIL (Python Imaging Library). In addition, the original implementation of faster-rcnn include a data augmentation 
\begin{itemize}
    \item{Brightness:\\
      with $factor = 1$ means no changes, we use $factor = [0.5,0.8,1.2,1.5]$
    }
    \item{Color: make the image colorful or not\\
      with $factor = 1$ means no changes, we use $factor = [0.5,0.8,1.2,1.5]$
    }
    \item{Contrast:\\
      with $factor = 1$ means no changes, we use $factor = [0.5,0.8,1.2,1.5]$
    }
    \item{Sharpness:\\
      with $factor = 1$ means no changes, we use $factor = [0.2,0.5,2.0,3.0]$
    }
\end{itemize}

In this way, we totally generated $4 * 4 * 4 * 4 * 1238 = 316928$ pictures, combined with the original picture, there are total about 20,000+ pictures for training (some of the 1238 original pictures are remained for testing, which could not be used to data augmentation)

Here are some demo of the original image and processed image:
\begin{figure}[htp]
    \centering
    \subfigure[Original Image]{\includegraphics[width=0.3\columnwidth]{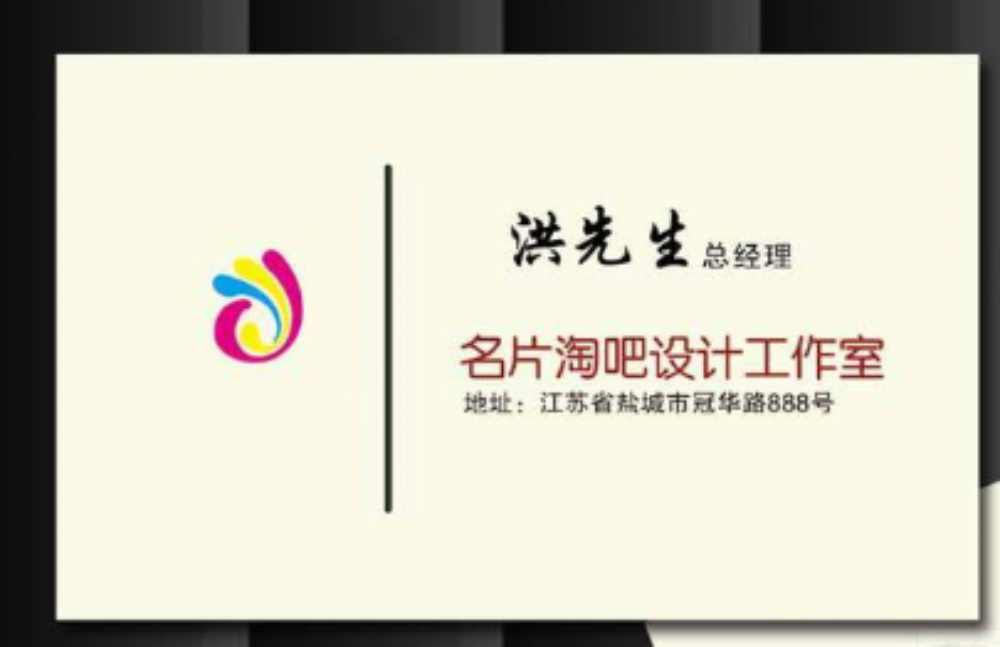}}
    \subfigure[Processed Image 1]{\includegraphics[width=0.3\columnwidth]{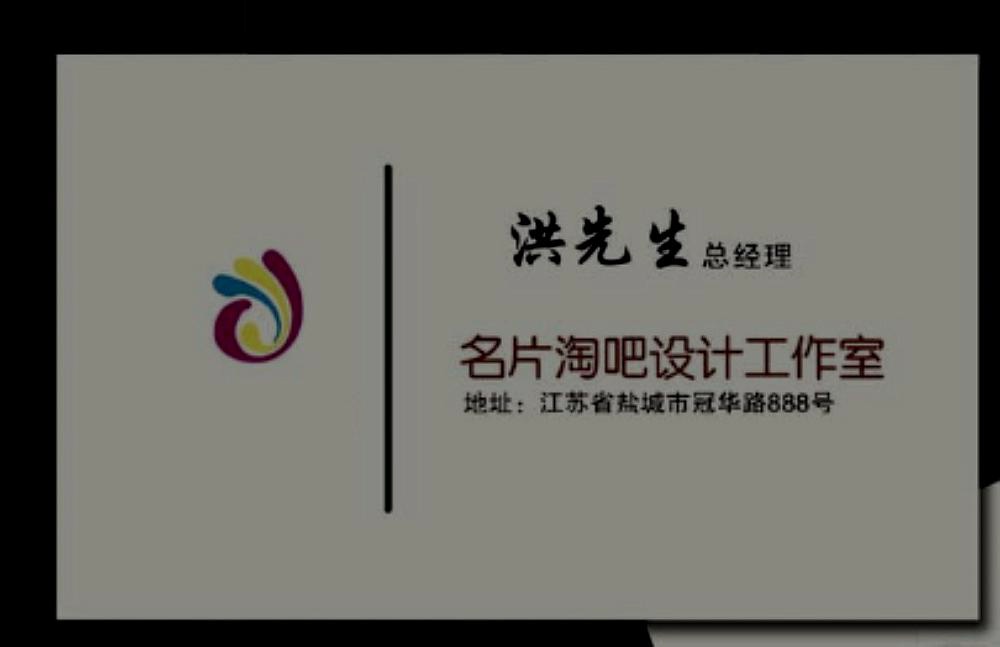}}
    \subfigure[Processed Image 2]{\includegraphics[width=0.3\columnwidth]{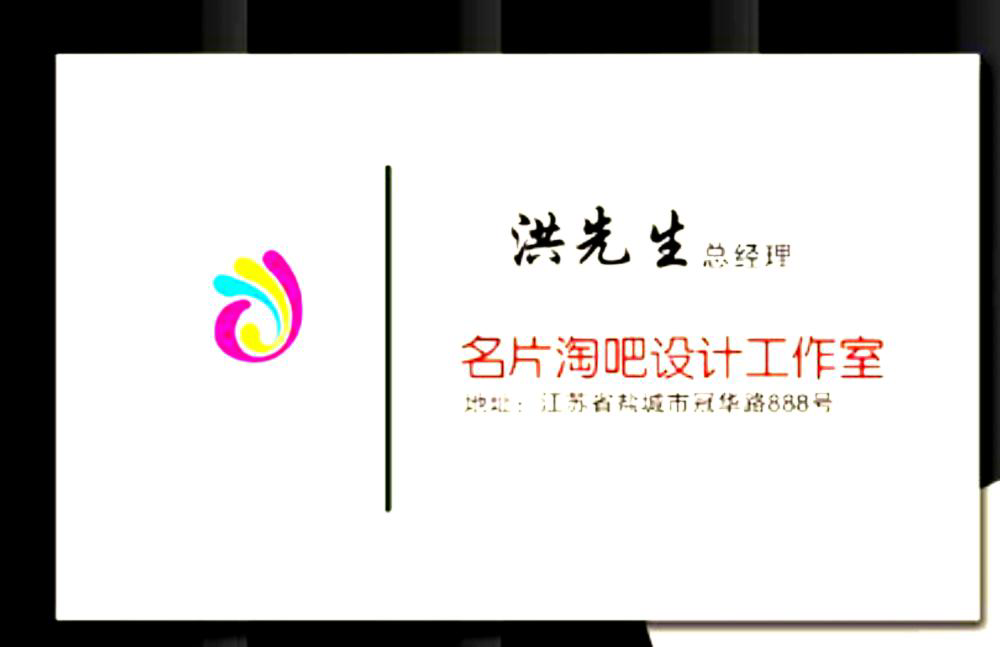}}
\end{figure}

\subsection{Fake Data Generation}
Besides the limit of data, we also found that the number of samples with the label 'English' and 'Number' is much fewer than the 
'Chinese' one. A convincing reason for this is that most of these namecard is collected in China, so, we have to generate some 'Fake'
namecard focus on English and Number to avoid the data inbalance. By putting some random English characters or numbers into some random
position of an image, we can generate some fake data just like the demo below.

However, we have to admit that the 'fake' data must have lost some information provided by the original namecard. Therefore, as a result,
the result will show that the addition of the 'fake' namecard will just improve the accuracy (i.e. mAP and IoU score) of English and 
Number about 2\%. This improvement might even be flooded by the noise of the model among different training random seed.

\begin{figure}[htp]
    \centering
    \subfigure[Fake Image 1]{\includegraphics[width=0.3\columnwidth]{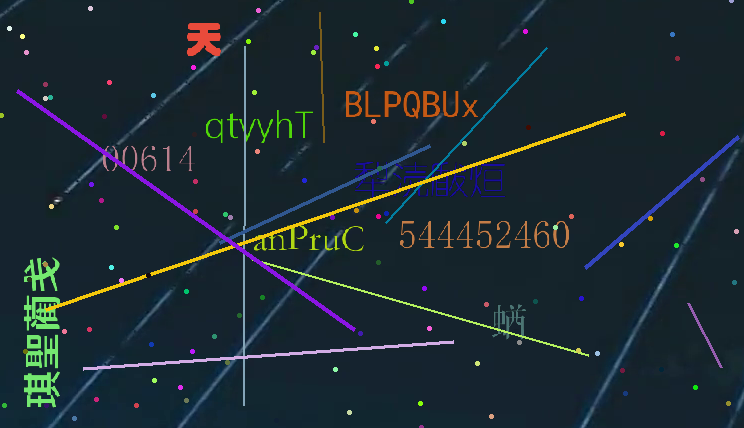}}
    \subfigure[Fake Image 2]{\includegraphics[width=0.3\columnwidth]{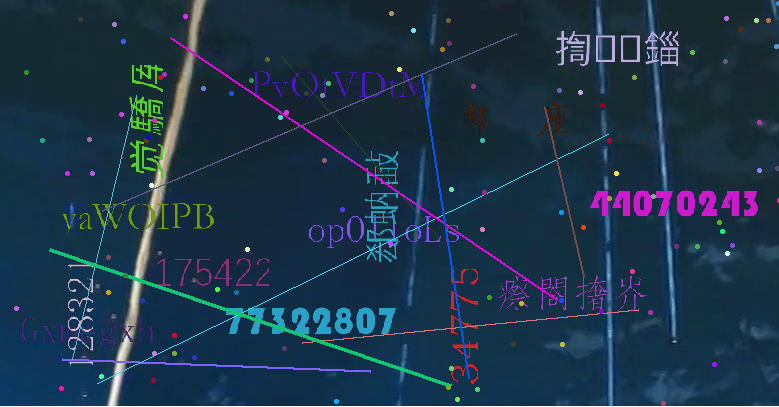}}
    \subfigure[Fake Image 3]{\includegraphics[width=0.3\columnwidth]{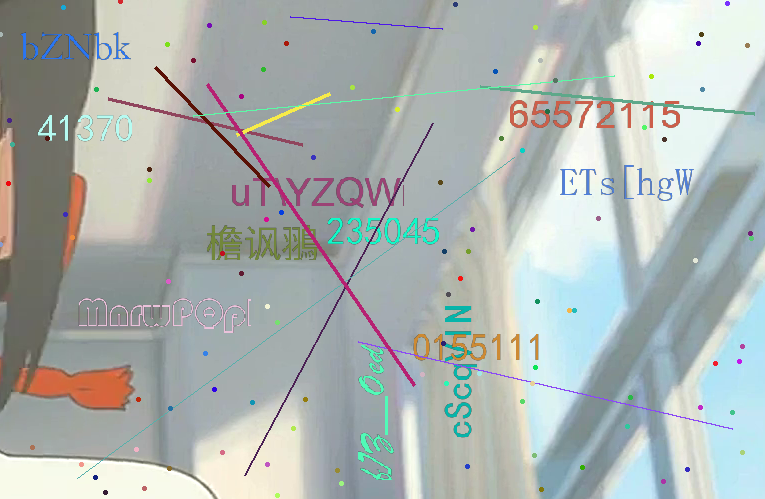}}
\end{figure}

\section{Experiments}
We build up the model with the help of Chen's previous work\cite{chen17implementation} on pyTorch. This version of 
faster RCNN is a little bit different from the original faster RCNN, however, all of the modifications would not 
affect the preformance a lot. We use the ImageNet pre-trained ResNet101 as our feature exactor.

\subsection{Traing and Testing on the original data set}
For 1225 training images, 14421 sample boxes, we train the network for 200,000 loops, each loop take about 0.5s on GPU, the total training time cost is about 1 day.

The score of IoU, mAP, F1 is presented as the following table, it is obvious that there is a little bit over-fitting for Chinese samples, however, over-fitting for English one and Number one is greater since that the English sample and Number sample are not as much as Chinese one.

\begin{table}[h]
    \centering\caption{Training Score using original data set}
    \begin{tabular}{ccccc}
        \toprule
        Score & Chinese  & English & Numer & Total \\
        \midrule
        IoU & 91.04\% & 89.86\% & 89.96\% & 90.47\%\\
        mAP & 94.50\% & 90.99\% & 93.16\% & 92.88\%\\
        F-1 & 96.51\% & 93.25\% & 95.37\% & 95.04\%\\
        \bottomrule
    \end{tabular}
\end{table}
\begin{table}[h]
    \centering\caption{Tesing Score using original data set}
    \begin{tabular}{ccccc}
        \toprule
        Score & Chinese  & English & Numer & Total \\
        \midrule
        IoU & 84.39\% &79.43\% & 81.81\% & 82.71\%\\
        mAP & 84.41\% & 63.21\% & 73.30\% & 73.64\%\\
        F-1 & 83.98\% & 63.65\% & 75.96\% & 74.53\%\\
        \bottomrule
    \end{tabular}
\end{table}

The training loss and the validation loss are shown below

\begin{figure}[h]
    \centering
    \includegraphics[width=\columnwidth]{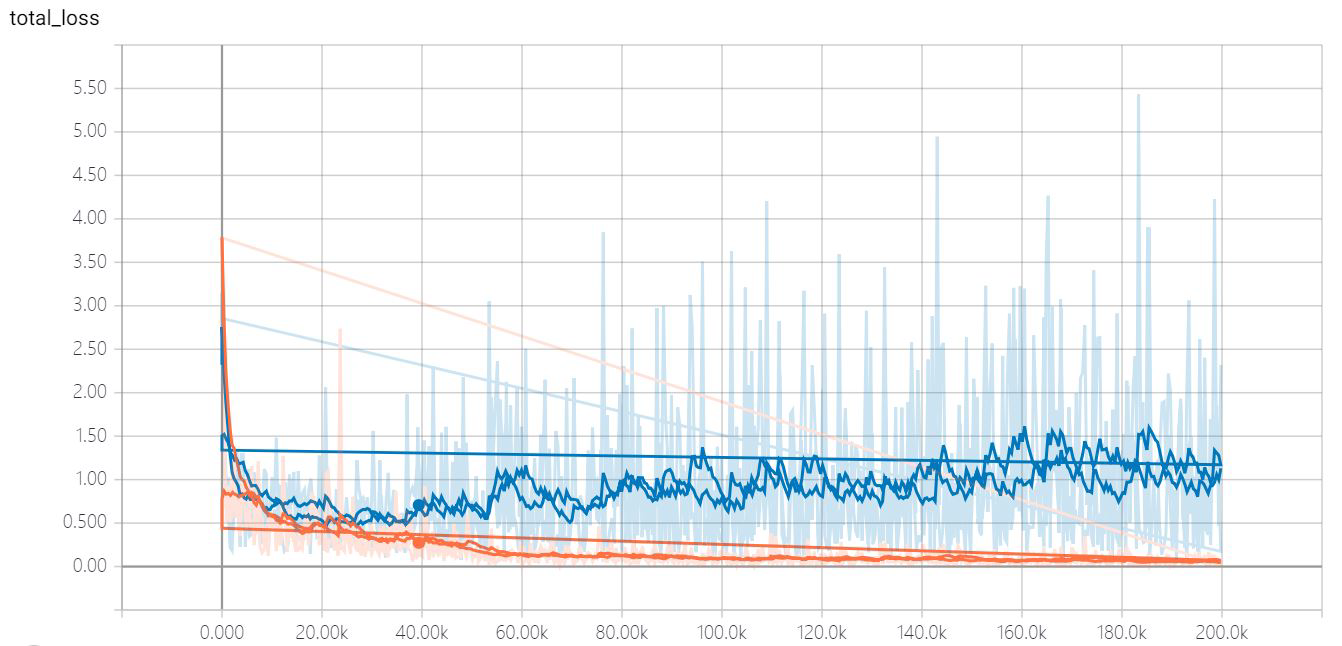}
    \caption{Training loss and the validation loss}
    \label{Ori}
\end{figure}

\subsection{Traing and Testing on the augmented data set}
For $1225*256 = 313,600$ training images, 3706,196 sample boxes, we train the network for 500,000 loops, each loop take about 0.6s on GPU, the total training time cost is about 3 day.

The score of IoU, mAP, F1 is presented as the following table, comparing to the origin dataset, the performance is improved a little, moreover, the overfitting of English and Number is fewer than the origin dataset. 

\begin{table}[h]
    \centering\caption{Training Score using original data set}
    \begin{tabular}{ccccc}
        \toprule
        Score & Chinese  & English & Numer & Total \\
        \midrule
        IoU & 90.31\% & 89.19\% & 89.73\% & 89.90\%\\
        mAP & 94.43\% & 91.41\% & 93.66\% & 93.16\%\\
        F-1 & 96.36\% & 93.36\% & 95.51\% & 95.08\%\\
        \bottomrule
    \end{tabular}
\end{table}
\begin{table}[h]
    \centering\caption{Tesing Score using original data set}
    \begin{tabular}{ccccc}
        \toprule
        Score & Chinese  & English & Numer & Total \\
        \midrule
        IoU & 86.48\% & 80.43\% & 82.91\% & 83.88\%\\
        mAP & 86.01\% & 64.92\% & 73.25\% & 74.72\%\\
        F-1 & 87.26\% & 65.73\% & 77.59\% & 75.86\%\\
        \bottomrule
    \end{tabular}
\end{table}

The training loss and the validation loss are shown below

\subsection{Against the model size limit: Using Half Float}

In order to struggle against the model size limit, we use the half float instead of float, i.e. using 16 digits float
instead of 32 digits float to save space. According to some experiments carried by Gupta et al. \cite{halffloat}. 
We can conclude that using half float in CNN network does not affect the precise of the network obviously

\subsection{Results}
More experiments shows that the 'fake namecard' generation would bring a little improvement to the preformance of 
the network, however, this improvement is so slight that it could be ignored. Moreover, we found out that this 
problem is kind of easy for the ResNet101 pretrained, using some light network such as mobileT might also bring 
good result.

We also find out that no matter what method is applied, there are still some problems with the English samples and Number samples, this might because the English Sample and the Number sample itself are different to distringuish, for example, 'B' and '8', 'O' and '0', 'l', 'I', and '1'...

Another possible reason is that the English charaters and the numbers are inserted easily to the Chinese samples and other samples, which makes it different to label.

Here are some output of the bounding box of name card, the Chinese characters are shown in red bounding box, 
the English one are showned in blue one, while the number is shown in black one

\begin{figure}[h]
    \centering
    \includegraphics[width=\columnwidth]{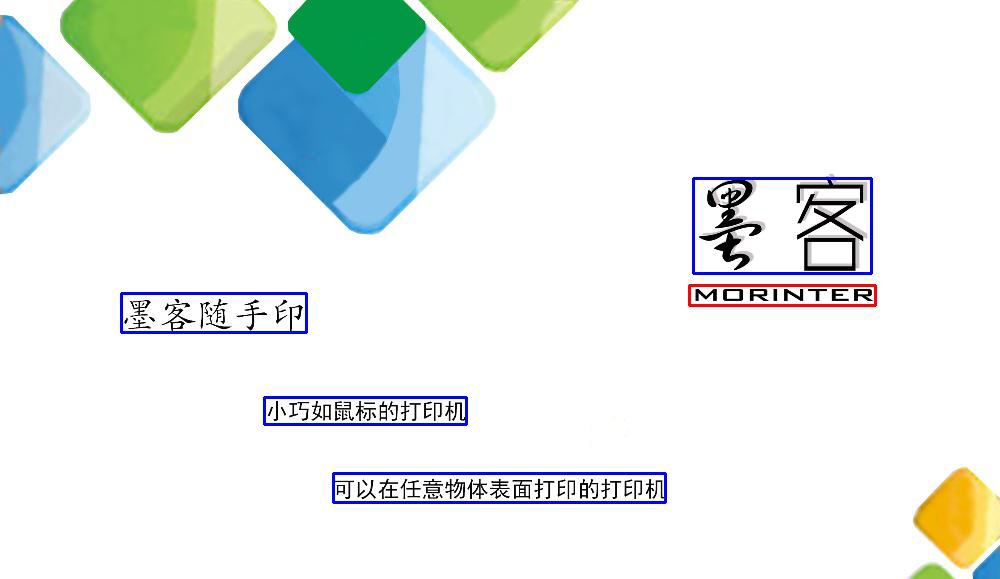}
    \caption{Image 0000}
\end{figure}

\begin{figure}[h]
    \centering
    \includegraphics[width=\columnwidth]{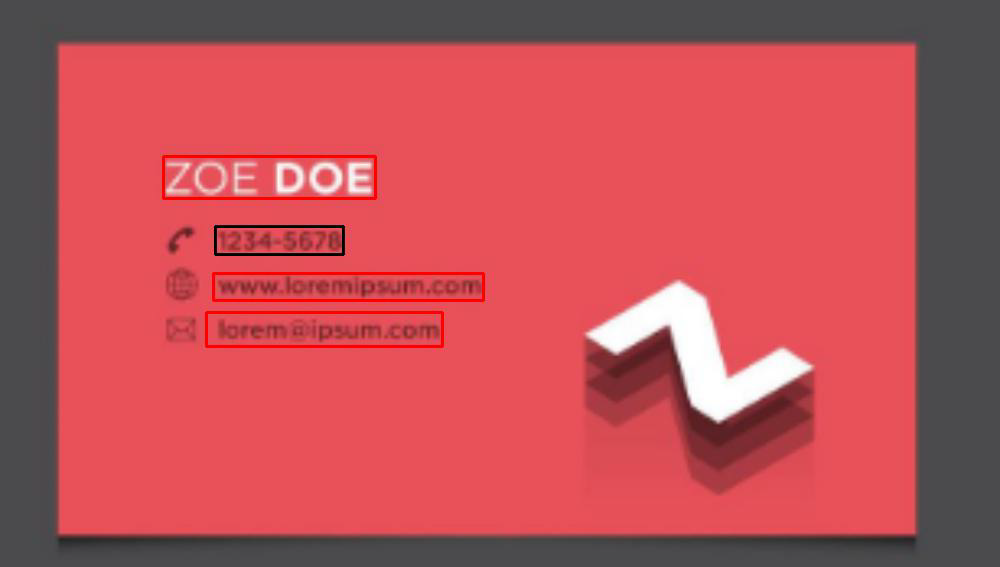}
    \caption{Image 0011}
\end{figure}

\section{Conclusion}

In conclusion, we found out that the faster RCNN model (with ResNet101 ImageNet pre-trained) can successfully detect the characters on namecard. And combined with other network structure, we can solve the detection and moreover, machine translation problems more easily.

\appendix{}
\subsection{Brief Introduction of Code}
Our code could be visited from \href
{https://github.com/ZeroWeight/Pattern-Recognize/tree/master/10_Name_Card_Detection}{\underline{here}}, below are some brief introduction of it.
\subsection{Prerequisites}
\begin{itemize}
\item{A basic pytorch 0.4 installation, CPython 3.5}

\item{Python packages you might not have: cffi, opencv-python, easydict 1.6 (similar to py-faster-rcnn).}

\item{tensorboard-pytorch to visualize the training and validation curve. Please build from source to use the latest tensorflow-tensorboard.}

\item{Setup: compile the NMS module and RoI-Pooling module: just run make.sh in ./pytorch-faster-rcnn/libs}
\end{itemize}

\subsection{Train}
We strongly recommend you not to train the network because it takes really a long time (more than 1 day)

\noindent
\ttfamily
\hlstd{}\hlkwb{cd\ }\hlstd{pytorch}\hlopt{{-}}\hlstd{faster}\hlopt{{-}}\hlstd{rcnn\hspace*{\fill}\\
.}\hlopt{/}\hlstd{experiments}\hlopt{/}\hlstd{scripts}\hlopt{/}\hlstd{train\textunderscore faster\textunderscore rcnn.sh\ }\hlkwd{\$GPU\textunderscore ID\ }\Righttorque\hspace*{\fill}\\
\hlstd{name\textunderscore card\textunderscore fake\ res101\ }\hlslc{\#\ for\ data{-}augmentation}\hspace*{\fill}\\
\hlstd{}\hlslc{\#\ OR}\hspace*{\fill}\\
\hlstd{.}\hlopt{/}\hlstd{experiments}\hlopt{/}\hlstd{scripts}\hlopt{/}\hlstd{train\textunderscore faster\textunderscore rcnn.sh\ }\hlkwd{\$GPU\textunderscore ID\ }\Righttorque\hspace*{\fill}\\
\hlstd{name\textunderscore card\textunderscore real\ res101\ }\hlslc{\#\ for\ data{-}augmentation}\hlstd{}\hspace*{\fill}\\
\mbox{}
\normalfont
\normalsize

\subsection{Test}

We have provided two testing method: batch mode or sample mode

\noindent
\ttfamily
\hlstd{}\hlkwb{cd\ }\hlstd{pytorch}\hlopt{{-}}\hlstd{faster}\hlopt{{-}}\hlstd{rcnn}\hlopt{/}\hlstd{tools}\hspace*{\fill}\\
\hspace*{\fill}\\
\hlslc{\#testing\ in\ batch}\hspace*{\fill}\\
\hlstd{CUDA\textunderscore VISIBLE\textunderscore DEVICES}\hlopt{=}\hlstd{}\hlkwd{\$GPU\textunderscore ID\ }\hlstd{python3.5\ }\hlkwb{test}\hlstd{.py\ namelist.txt}\hspace*{\fill}\\
\hspace*{\fill}\\
\hlslc{\#testing\ per\ sample}\hspace*{\fill}\\
\hlstd{CUDA\textunderscore VISIBLE\textunderscore DEVICES}\hlopt{=}\hlstd{}\hlkwd{\$GPU\textunderscore ID\ }\hlstd{python3.5\ }\hlkwb{test}\hlstd{.py\ }\hlkwb{test}\hlstd{.jpg}\hspace*{\fill}\\
\mbox{}
\normalfont
\normalsize

\subsubsection{Sample Mode}
By inputing a single .jpg file, you have chosen the sample mode, the predict bounding box are output to the 
standard output, a new jpg file with the same name with the original one are generated into the ./pytorch-faster-rcnn/tools. Where blue bounding box stands for Chinese, red one stands for English while the black bounding box stands for Number.

Using stream redirect to keep the output in log

\subsection{Batch Mode}
If you would like to test a batch of images, we recommend you to use the batch mode, just by inputing a .txt file, where each line is a path to the image (relative path from ./pytorch-faster-rcnn/tools or full path). Output and Output Images are the same with sample mode, however, it is more faster since it loads the model for only one time.

\section*{acknowledgement}

I would like to express my deep gratitude to Professor ZHANG, my research supervisors, for his patient guidance on the basic knowledge of pattern recognition, enthusiastic encouragement and useful critiques of this research work. I would also like to thank all of my classmates and parteners, for their warm help and assisstance in collecting the namecard data. My grateful thanks are also extended to Mr. Lu for his help in help me dealing with some problems of this project.

I would also like to extend my thanks to the technicians of the laboratory of the Department of Automation, Tsina University for their help in offering me the server and GPU resources in running the program.

Finally, I wish to thank my parents for their support and encouragement throughout my study.

\bibliographystyle{ieeetran}
\bibliography{reference}

\begin{thebibliography}{10}
\providecommand{\url}[1]{#1}
\csname url@samestyle\endcsname
\providecommand{\newblock}{\relax}
\providecommand{\bibinfo}[2]{#2}
\providecommand{\BIBentrySTDinterwordspacing}{\spaceskip=0pt\relax}
\providecommand{\BIBentryALTinterwordstretchfactor}{4}
\providecommand{\BIBentryALTinterwordspacing}{\spaceskip=\fontdimen2\font plus
\BIBentryALTinterwordstretchfactor\fontdimen3\font minus
  \fontdimen4\font\relax}
\providecommand{\BIBforeignlanguage}[2]{{%
\expandafter\ifx\csname l@#1\endcsname\relax
\typeout{** WARNING: IEEEtran.bst: No hyphenation pattern has been}%
\typeout{** loaded for the language `#1'. Using the pattern for}%
\typeout{** the default language instead.}%
\else
\language=\csname l@#1\endcsname
\fi
#2}}
\providecommand{\BIBdecl}{\relax}
\BIBdecl

\bibitem{GehringAGYD17}
\BIBentryALTinterwordspacing
J.~Gehring, M.~Auli, D.~Grangier, D.~Yarats, and Y.~N. Dauphin, ``Convolutional
  sequence to sequence learning,'' \emph{CoRR}, vol. abs/1705.03122, 2017.
  [Online]. Available: \url{http://arxiv.org/abs/1705.03122}
\BIBentrySTDinterwordspacing

\bibitem{VaswaniSPUJGKP17}
\BIBentryALTinterwordspacing
A.~Vaswani, N.~Shazeer, N.~Parmar, J.~Uszkoreit, L.~Jones, A.~N. Gomez,
  L.~Kaiser, and I.~Polosukhin, ``Attention is all you need,'' \emph{CoRR},
  vol. abs/1706.03762, 2017. [Online]. Available:
  \url{http://arxiv.org/abs/1706.03762}
\BIBentrySTDinterwordspacing

\bibitem{DBLP:journals/corr/abs-1709-02755}
\BIBentryALTinterwordspacing
T.~Lei, Y.~Zhang, and Y.~Artzi, ``Training rnns as fast as cnns,'' \emph{CoRR},
  vol. abs/1709.02755, 2017. [Online]. Available:
  \url{http://arxiv.org/abs/1709.02755}
\BIBentrySTDinterwordspacing

\bibitem{DBLP:journals/corr/GuCL17}
\BIBentryALTinterwordspacing
J.~Gu, K.~Cho, and V.~O.~K. Li, ``Trainable greedy decoding for neural machine
  translation,'' \emph{CoRR}, vol. abs/1702.02429, 2017. [Online]. Available:
  \url{http://arxiv.org/abs/1702.02429}
\BIBentrySTDinterwordspacing

\bibitem{renNIPS15fasterrcnn}
S.~Ren, K.~He, R.~Girshick, and J.~Sun, ``Faster {R-CNN}: Towards real-time
  object detection with region proposal networks,'' in \emph{Advances in Neural
  Information Processing Systems ({NIPS})}, 2015.

\bibitem{ILSVRC15}
O.~Russakovsky, J.~Deng, H.~Su, J.~Krause, S.~Satheesh, S.~Ma, Z.~Huang,
  A.~Karpathy, A.~Khosla, M.~Bernstein, A.~C. Berg, and L.~Fei-Fei, ``Imagenet
  large scale visual recognition challenge,'' \emph{International Journal of
  Computer Vision (IJCV)}, vol. 115, no.~3, pp. 211--252, 2015.

\bibitem{VGG}
K.~Simonyan and A.~Zisserman, ``Very deep convolutional networks for
  large-scale image recognition,'' \emph{arXiv preprint arXiv:1409.1556}, 2014.

\bibitem{RES}
K.~He, X.~Zhang, S.~Ren, and J.~Sun, ``Deep residual learning for image
  recognition,'' in \emph{Proceedings of the IEEE conference on computer vision
  and pattern recognition}, 2016, pp. 770--778.

\bibitem{girshickICCV15fastrcnn}
R.~Girshick, ``Fast r-cnn,'' in \emph{International Conference on Computer
  Vision ({ICCV})}, 2015.

\bibitem{chen17implementation}
X.~Chen and A.~Gupta, ``An implementation of faster rcnn with study for region
  sampling,'' \emph{arXiv preprint arXiv:1702.02138}, 2017.

\bibitem{halffloat}
\BIBentryALTinterwordspacing
S.~Gupta, A.~Agrawal, K.~Gopalakrishnan, and P.~Narayanan, ``Deep learning with
  limited numerical precision,'' \emph{CoRR}, vol. abs/1502.02551, 2015.
  [Online]. Available: \url{http://arxiv.org/abs/1502.02551}
\BIBentrySTDinterwordspacing

\end{thebibliography}
\end{document}